\documentclass[11pt,a4paper]{article}
\usepackage[hyperref]{naaclhlt2019}
\usepackage{times} 
\usepackage{latexsym}

\usepackage{booktabs}
\usepackage{multirow}
\usepackage{amsmath}
\usepackage{url}
\usepackage{graphicx}
\usepackage{helvet}  
\usepackage{courier}
\usepackage{threeparttable}
\def\tran{^\mathrm{\scriptscriptstyle T}}
\aclfinalcopy 

\newcommand{\tabincell}[2]{\begin{tabular}{@{}#1@{}}#2\end{tabular}}

\usepackage{CJKutf8}

\title{VCWE: Visual Character-Enhanced Word Embeddings}

\author{Chi Sun, Xipeng Qiu\thanks{{ }{ }Corresponding author.}, Xuanjing Huang\\
 Shanghai Key Laboratory of Intelligent Information Processing, Fudan University\\
School of Computer Science, Fudan University\\
825 Zhangheng Road, Shanghai, China\\
{\tt \{sunc17,xpqiu,xjhuang\}@fudan.edu.cn} \\}

\date{}

\begin{document}
\begin{CJK*}{UTF8}{gbsn}
		\maketitle
		\begin{abstract}
			Chinese is a logographic writing system, and the shape of Chinese characters contain rich syntactic and semantic information. In this paper, we propose a model to learn Chinese word embeddings via three-level composition: (1) a convolutional neural network to extract the intra-character compositionality from the visual shape of a character; (2) a recurrent neural network with self-attention to compose character representation into word embeddings; (3) the Skip-Gram framework to capture non-compositionality directly from the contextual information. Evaluations demonstrate the superior performance of our model on four tasks: word similarity, sentiment analysis, named entity recognition and part-of-speech tagging.\footnote{The source codes are available at \url{https://github.com/HSLCY/VCWE}}
		\end{abstract}
		\section{Introduction}

		Distributed representations of words, namely word embeddings, encode both semantic and syntactic information into a dense vector.
		Currently, word embeddings have been playing a pivotal role in many natural language processing (NLP) tasks. Most of these NLP tasks also benefit from the pre-trained word embeddings, such as \emph{word2vec}~\cite{mikolov2013efficient} and GloVe~\cite{pennington2014glove}, which are based on the distributional hypothesis \cite{harris1954distributional}: words that occur in the same contexts tend to have similar meanings.
		Earlier word embeddings often take a word as a basic unit, and they ignore compositionality of its sub-word information such as morphemes and character n-grams, and cannot competently handle the rare words.
		To improve the performance of word embeddings, sub-word information has been employed \cite{luong2013better,qiu2014co,cao2016joint,sun2016inside,wieting2016charagram,bojanowski2016enriching}.

		Compositionality is more critical for Chinese, since
		Chinese is a logographic writing system. In Chinese, each word typically consists of fewer characters and each character also contains richer semantic information. For example, Chinese character ``休'' (rest) is composed of the characters for ``人'' (person) and ``木'' (tree), with the intended idea of someone leaning against a tree, i.e., resting.
		
		Based on the linguistic features of Chinese, recent methods have used the character information to improve Chinese word embeddings. These methods can be categorized into two kinds:
		
		1) One kind of methods learn word embeddings with its constituent character~\cite{chen2015joint}, radical\footnote{the graphical component of Chinese, referring to \url{https://en.wikipedia.org/wiki/Radical_(Chinese_characters)}}
		~\cite{shi2015radical,yin2016multi,yu2017joint} or strokes\footnote{the basic pattern of Chinese characters, referring to \url{https://en.wikipedia.org/wiki/Stroke_(CJKV_character)}}~\cite{cao2018cw2vec}.
		However, these methods usually use simple operations, such as averaging and n-gram, to model the inherent compositionality within a word, which is not enough to handle the complicated linguistic compositionality.
		
		2) The other kind of methods learns word embeddings with the visual information of the character. \citet{liu2017learning} learn character embedding based on its visual characteristics in the text classification task.
		\citet{su2017learning} also introduce a pixel-based model that learns character features from font images. However, their model is not shown to be better than \emph{word2vec} model because it has little flexibility and fixed character features.
		
		Besides, most of these methods pay less attention to the non-compositionality. For example, the semantic of Chinese word ``沙发'' (sofa) cannot be composed by its contained characters ``沙'' (sand) and ``发'' (hair).
		
		In this paper, we fully consider the compositionality and non-compositionality of Chinese words and propose a visual character-enhanced word embedding model (VCWE) to learn Chinese word embeddings. VCWE learns Chinese word embeddings via three-level composition:
		\begin{itemize}
		\item The first level is to learn the intra-character composition, which gains the representation of each character from its visual appearance via a convolutional neural network;
		
		\item The second level is to learn the inter-character composition, where a bidirectional long short-term neural network (Bi-LSTM) \cite{hochreiter1997long} with self-attention to compose character representation into word embeddings;
		
		\item The third level is to learn the non-compositionality, we can learn the contextual information because the overall framework of our model is based on the skip-gram.
		\end{itemize}
		Evaluations demonstrate the superior performance of our model on four tasks such as word similarity, sentiment analysis, named entity recognition and part-of-speech tagging.
		
		\section{Related Work}
		In the past decade, there has been much research on word embeddings. \citeauthor{bengio2003neural} \shortcite{bengio2003neural} use a feedforward neural network language model to predict the next word given its history. Later methods~\cite{mikolov2010recurrent} replace feedforward neural network with the recurrent neural network for further exploration. The most popular word embedding system is \emph{word2vec}, which uses continuous-bag-of-words and Skip-gram models, in conjunction with negative sampling for efficient conditional probability estimation \cite{mikolov2013efficient}.
		
		A different way to learn word embeddings is through factorization of word co-occurrence matrices such as GloVe embeddings \cite{pennington2014glove}, which have been shown to be intrinsically linked to Skip-gram and negative sampling \cite{levy2014neural}.
		
		The models mentioned above are popular and useful, but they regard individual words as atomic tokens, and the potentially useful internal structured information of words is ignored. To improve the performance of word embedding, sub-word information has been employed \cite{luong2013better,qiu2014co,cao2016joint,sun2016inside,wieting2016charagram,bojanowski2016enriching}. These methods focus on alphabetic writing systems, but they are not directly applicable to logographic writing systems.
		
		For the alphabetic writing systems, research on Chinese word embedding has gradually emerged. These methods focus on the discovery of making full use of sub-word information. \citeauthor{chen2015joint} \shortcite{chen2015joint} design a CWE model for jointly learning Chinese characters and word embeddings. Based on the CWE model, \citeauthor{yin2016multi} \shortcite{yin2016multi} present a multi-granularity embedding (MGE) model, additionally using the embeddings associated with radicals detected in the target word. \citet{xu2016improve} propose a similarity-based character-enhanced word embedding (SCWE) model, exploiting the similarity between a word and its component characters with the semantic knowledge obtained from other languages. \citeauthor{shi2015radical} \shortcite{shi2015radical} utilize radical information to improve Chinese word embeddings. \citeauthor{yu2017joint} \shortcite{yu2017joint} introduce a joint learning word embedding (JWE) model and \citeauthor{cao2018cw2vec} \shortcite{cao2018cw2vec} represent Chinese words as sequences of strokes and learn word embeddings with stroke n-grams information.
		
		From another perspective, \citeauthor{liu2017learning} \shortcite{liu2017learning} provide a new way to automatically extract character-level features, creating an image for the character and running it through a convolutional neural network to produce a visual character embedding.  \citeauthor{su2017learning} \shortcite{su2017learning} also introduce a pixel-based model that learns character features from its image.
		
		Chinese word embeddings have recently begun to be explored, and have so far shown great promise. In this paper, we propose a visual character-enhanced word embedding (VCWE) model that can learn Chinese word embeddings from corpus and images of characters. The model combines the semantic information of the context with the image features of the character, with superior performance in several benchmarks.

		\section{Proposed Model}
		
		\begin{figure*}[t]
			\center{\includegraphics[width=17cm]  {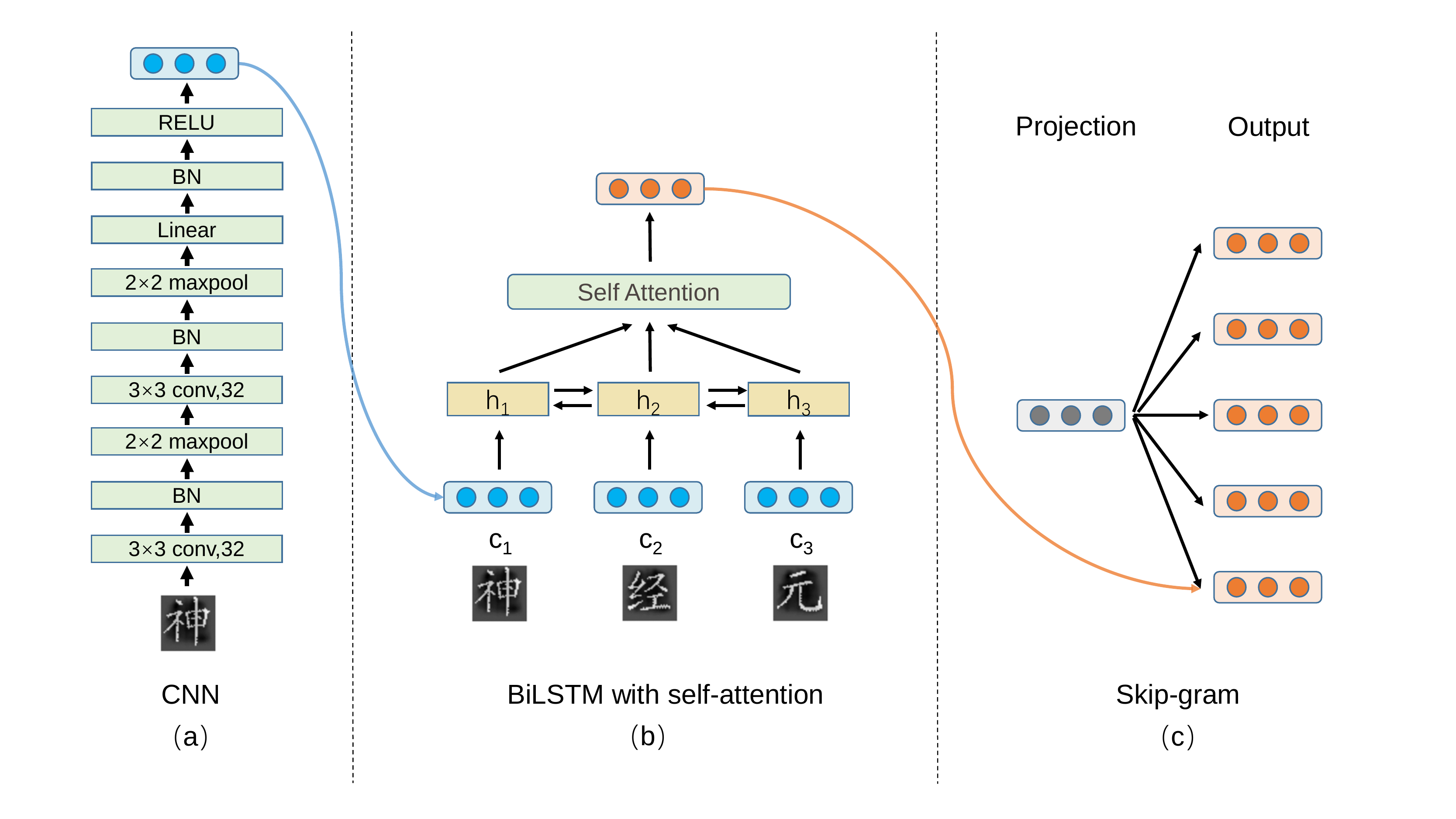}}
			\caption{\label{fig_1} The overall architecture of our approach.}
		\end{figure*}
		In this section, we introduce the visual character-enhanced word embedding (VCWE) model for Chinese word representation.
		
		Given a Chinese word $w$ consisting of $n$ characters $c_1,\cdots,c_n$, its semantic may come from either its contained characters or its contexts. Therefore, we use the two-level hierarchical composition to compose the word embedding, which further learned according to its context.
		
		The overall architecture of our approach is on Figure \ref{fig_1}. We first use a convolutional neural network (CNN) to model the intra-character compositionality of character from its visual shape information. We use the output of CNN as the embeddings of the character. Then the character embeddings are used as the input of the bidirectional LSTM network to model the inter-character compositionality. After a self-attention layer, we can get the representation of the word. Finally, based on the Skip-Gram framework, we learn the word embeddings with the visual character-enhanced embedding of the context.
		
		\subsection{Intra-Character Composition}
		
		Since the shape of a Chinese character provides rich syntactic and semantic information, the representation of a character can be composed by its intrinsic visual components.
		Following the success of the convolutional neural network (CNN) \cite{lecun1995convolutional} in computer vision, we use CNN to directly model the natural composition of a character from its image.
		
		We first convert each character into an image of size $40\times 40$, a deep CNN is used to fuse its visual information fully. The specific structure of the CNN is shown in Figure \ref{fig_1}(a), which consists of two convolution layers and one linear layer. Each convolution layer is followed by a max pooling layer and a batch normalization layer. The lower layers aim to capture the stroke-level information, and the higher layers aim to capture the radical-level and component-level information.
		
		The output of CNN can be regarded as the representation of the character. The character representation by its visual information can fully capture its intrinsic syntactic and semantic information with the intra-character compositionality.
		
		The parameters of CNN are learned through backpropagation in end-to-end fashion.

		\subsection{Inter-Character Composition}
		
		After obtaining the representation of characters, we combine them into word embedding.
		The word embedding need to capture the character-level compositionality fully. Here, we use the bidirectional LSTM (Bi-LSTM) \cite{hochreiter1997long} with self-attention to fuse the inter-character information of a word.
		
		The structure of our Bi-LSTM with self-attention is shown in Figure \ref{fig_1}(b).
		
		Given a word $w$ consisting of $n$ characters $c_1,\cdots,c_n$, we use $\mathbf{e}_1,\cdots,\mathbf{e}_n$ denote is the character representations, which are the output of the CNN rather than randomly initialized.
		
		The word $w$ is firstly encoded using a Bi-LSTM:
		\begin{align} \mathbf{h}_i^{F}&=\mathrm{LSTM}(\mathbf{h}_{i-1}^{F},\mathbf{e}_i),\\
		\mathbf{h}_i^{B}&=\mathrm{LSTM}(\mathbf{h}_{i+1}^{B},\mathbf{e}_i),\\
		\mathbf{h}_i &=[\mathbf{h}_i^{F};\mathbf{h}_i^{B}],\\
		H&=[\mathbf{h}_1,\mathbf{h}_2,...,\mathbf{h}_n],
		\end{align}
		where $\mathbf{h}_i$ is the hidden state of the $i$-th character in $w$.
		
		Then we use self-attention to obtain the inter-character compositionality.
		Following the self-attention proposed by \cite{lin2017structured}, we compute an attention vector $\alpha$:
		\begin{align}
		\alpha=\mathrm{softmax}(\mathbf{v}\tanh(U\mathbf{h}_{i}\tran)),
		\end{align}
		where $\mathbf{v}$ and $U$ are learnable weight parameters.
		
		Finally, the representation of word $w$ is:
		\begin{align}
		\mathbf{m}=\sum_{i=1}^{n}\alpha_i \mathbf{h}_i.
		\end{align}
		
		Since the Bi-LSTM's hidden state of each character is different according to its contexts, we believe the hidden state can capture both the compositional and non-compositional relations of the characters within a word.
		
		After obtaining the word representation, Skip-Gram \cite{mikolov2013efficient} is used to learn the word embedding with its context information.
		Skip-Gram is a useful framework for learning word vectors, which aims to predict context words given a target word in a sentence.
		
		Given a pair of words $(w, c)$, we denote $p(c|w)$ as the probability that the word $c$ is observed in the context of the target word $w$.
		
		With the negative-sampling approach, skip-gram formulates the probability $p(c|w)$ as follows:
		
		Given a pair of words $(w, c)$, the probability that the word $c$ is observed in the context of the target word $w$ is given by
		
		\begin{align}
		p(D = 1|w, c) &= \sigma(\mathbf{w}\tran \mathbf{c}),
		\end{align}
		where $\mathbf{w}$ and $\mathbf{c}$ are embedding vectors of $w$ and $c$ respectively, $\sigma$ is the sigmoid function.
		
		The probability of not observing word $c$ in the context of $w$ is given by:
		\begin{align}
		p(D = 0|w, c) &= 1 - \sigma(\mathbf{w}\tran \mathbf{c}).
		\end{align}
		
		\section{Training}
		
		\subsection{Objective Function}
		
		Given the target word $w$, its context word $c$ and $k$ negative words $\tilde{c}_1,...,\tilde{c}_k$. The word $w$ is a word selected from a sentence in the corpus, and the context $c$ is a nearby word within a window size $l$. The negative sample $\tilde{c}_i$ is a word that is randomly sampled at a certain frequency in the vocabulary.
		
		The loss function of VCWE model is as follows:
		\begin{align}
		L=L_1&+ L_2,\\
		L_1=\log \sigma(\mathbf{w}\tran\mathbf{c}) +&\sum_{i=1}^{k} \log \sigma(-\mathbf{w}\tran\tilde{\mathbf{c}}_{i}),\\
		L_2=\log \sigma(\mathbf{w}\tran\mathbf{m}_c) + & \sum_{i=1}^{k} \log \sigma(-\mathbf{w}\tran\tilde{\mathbf{m}}_{i}),
		\end{align}
		where 
		$\mathbf{w}$ is the lookup embedding of target word;
		$\mathbf{c}$ and $\tilde{\mathbf{c}}_{i}$ are the lookup embeddings of the context and negative words respectively; $\mathbf{m}_c$ and $\tilde{\mathbf{m}}_{i}$ are visual enhanced word embeddings of the context and negative words respectively.
		
		Here, we use the visually enhanced word embedding as the representation of context word instead of the target word. The final embedding of the target word is indirectly affected by the visual information. Thus, the final word embedding can have an advantage of fully utilizing intra-character compositionality from CNN, inter-character compositionality from LSTM, and context information from Skip-gram.
		
		\subsection{Word Sampling}
		We use a word sampling scheme similar to the implementation in
		word2vec \cite{mikolov2013efficient,mikolov2013distributed} to balance the importance of frequent words and rare words. Frequent words such as ``的''(of), ``是''(is), ``这''(this) are not as meaningful as relatively less frequent words such as ``猫''(cat), ``喜欢''(like), ``水果''(fruit). To improve the performance of word embeddings, we use subsampling\cite{mikolov2013distributed} to discard the word $w$ with the probability of $P(w)=1-\sqrt{\frac{t}{f(w)}}$ when generating the batch, where $f(w)$ is the frequency of word $w$ and $t$ is a chosen threshold, typically around $10^{-5}$.
		
		To generate negative context words, we sample each word $w$ according to distribution $P(w)\propto U(w)^{\frac{3}{4}}$, where $U(w)$ is the unigram distribution, which is the frequency of single words appearing in the corpus. This method also plays a role in reducing the frequency of occurrence of high-frequency words.

		\section{Experiments}
		\subsection{Preparation for training data}
		We download Chinese Wikipedia dump\footnote{https://dumps.wikimedia.org/zhwiki/20180520/} on May 20, 2018, which consists of 278K Chinese Wikipedia articles. We use the WikiExtractor toolkit\footnote{https://github.com/attardi/wikiextractor/blob/master/Wiki\\Extractor.py} to convert data from XML into text format. We find that the corpus consists of both simplified and traditional Chinese characters. Hence we utilize the opencc toolkit\footnote{https://github.com/BYVoid/OpenCC} to normalize all characters as simplified Chinese. We remove non-Chinese characters such as punctuation marks by retaining the characters whose Unicode falls into the range between 0x4E00 and 0x9FA5. We use THULAC  \footnote{https://github.com/thunlp/THULAC-Python}\cite{sun2016thulac} for word segmentation.
		
		We discard words that appeared less than 100 times and obtain a vocabulary of size 66,856. We count the frequency of occurrence of each word to prepare for the subsampling work.
		
\begin{figure}[htb]
			\center{\includegraphics[width=8cm]  {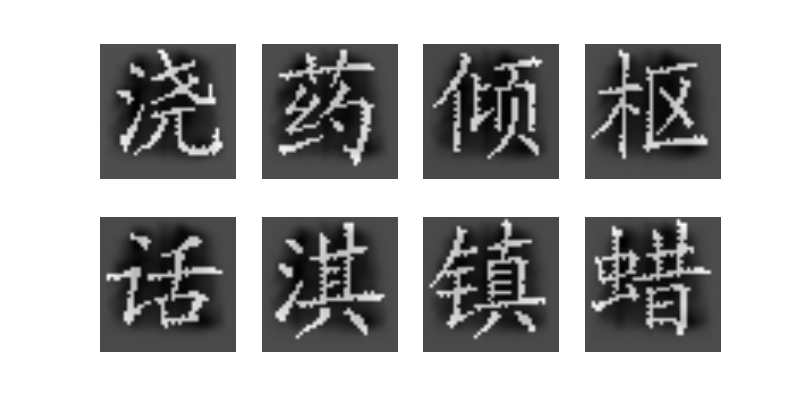}}
			\caption{\label{fig_2} The pre-processed Chinese character images.}
		\end{figure}

		In all 66,856 words, we extract 5030 unique characters. We use a Chinese character image generation software to generate the images of these Chinese characters. We subtract a mean image from each input image to center it before feeding it into the CNN. The pre-processed Chinese character images are shown in Figure \ref{fig_2}.
		
		\begin{table*}[t!]
			\centering
			\begin{tabular}{l|c c c c|c c}
				\toprule
				Model & WS-240 & WS-296 &  MC-30 & RG-65 & avg & $\Delta$ \\
				\midrule
				Skip-gram & 50.23 & 56.94 & 69.66 & 59.86 & 59.17 & - \\
				CBOW & 51.49 & 61.01 & 68.97 & 63.85 & 61.33 & +2.16 \\
				CWE & 52.63 & 58.98 & 68.82 & 59.60 & 60.01 & +0.84\\
				GWE & 52.74 & 58.22 & 68.23 & 60.74 & 59.98 & +0.81\\
				JWE & 51.92 & 59.84 & 70.27 & 62.83 & 61.22 & +2.05\\
				\midrule
				VCWE & 57.81 & \textbf{61.29} & \textbf{72.77} & \textbf{70.62} & \textbf{65.62} & \textbf{+6.45}\\
				\quad -CNN & 55.82 & 59.60 & 66.87 & 68.53 & 62.71 & +3.54\\
				\quad -LSTM & \textbf{58.13} & 60.85 & 68.03 & 69.78 & 64.20 & +5.03\\
				\bottomrule
			\end{tabular}
			\caption{\label{table_1}Spearman correlation for word similarity datasets, ``-CNN'' represents replacing the CNN and image information with randomly initialized character embedding, ``-LSTM'' represents replacing Bi-LSTM network and self-attention with the averaging operation. For each dataset, we boldface the score with the best performance across all models.
			}
		\end{table*}
		\begin{table*}[t!]
			\centering
			\begin{tabular}{l|c c c c c|c c}
				\toprule
				Model & NOTEBOOK & CAR &  CAMERA & PHONE & ALL & avg & $\Delta$ \\
				\midrule
				Skip-gram & 69.84 & 77.12 & 80.80 & 81.25 & 86.65 & 79.13 & - \\
				CBOW & 74.60 & 75.42 & 82.59 & 82.81 & 84.07 & 79.90 & +0.77 \\
				CWE & 73.02 & 80.51 & 81.25 & 81.25 & 82.09 & 79.62 & +0.49 \\
				GWE & 74.60 & 78.81 & 79.46 & 83.98 & 83.92 & 80.15 & +1.02 \\
				JWE & 77.78 & 78.81 & 81.70 & 81.64 & 85.13 & 81.01 & +1.88 \\
				\midrule
				VCWE & 80.95 & \textbf{85.59} & \textbf{83.93} & \textbf{84.38} & \textbf{88.92} & \textbf{84.75} & \textbf{+5.62} \\
				\quad -CNN  & \textbf{84.13} & 81.36 & 81.70 & 83.69 & 84.22 & 83.02 & +3.89\\
				\quad -LSTM & 79.37 & 80.51 & 80.36 & \textbf{84.38} & 85.58 & 82.04 & +2.91\\
				\bottomrule
			\end{tabular}
			\caption{\label{table_2}Accuracy for Sentiment analysis task. The configurations are the same of the ones used in Table \ref{table_1}.
			}
		\end{table*}
		\subsection{Hyperparameters}
		Models used for evaluation have dimension
		$D = 100$ and use context window $l = 5$ unless
		stated otherwise. We use the threshold $t = 10^{-5}$ for subsampling, which is the recommended value for word2vec Skip-gram \cite{mikolov2013efficient} on large datasets. The number of negative samples per word is 5.
		
		We use mini-batch asynchronous gradient descent with Adam \cite{kingma2014adam}. The initial learning rate is 0.001.
		
		\subsection{Baselines}
		We compare our model to the following open source state-of-art models:
		\begin{itemize}
			\item word2vec\footnote{https://code.google.com/archive/p/word2vec/} \cite{mikolov2013efficient} is arguably the most popular word embedding, which uses continuous-bag-of-words (CBOW) and Skip-gram models. We train word2vec with both Skip-gram and CBOW models. We did not train Glove\cite{pennington2014glove} because it did not perform well in many previous Chinese word embedding papers.
			\item
			CWE\footnote{https://github.com/Leonard-Xu/CWE} \cite{chen2015joint} is character-enhanced word embeddings which introduce internal character information into word embedding methods to alleviate excessive reliance on the external information.
			\item
			GWE\footnote{https://github.com/ray1007/gwe}\cite{su2017learning}
			is a pixel-based Chinese word embedding model, which exploits character features from font images by convolutional autoencoders.
			\item
			JWE\footnote{https://github.com/hkust-knowcomp/jwe}\cite{yu2017joint} is a model to jointly learn the embeddings of Chinese words, characters, and sub character components.
		\end{itemize}
		
		For a fair comparison between different algorithms, we use the same corpus and the same hyperparameters mentioned in previous subsections.
		
		\subsection{Word Similarity Task}
		We evaluate our embeddings on the Chinese
		word similarity datasets wordsim-240 and wordsim-296 provided by \cite{chen2015joint}. Besides, we translate two English word similarity datasets MC-30 \cite{miller1991contextual} and RG-65 \cite{rubenstein1965contextual} to Chinese\footnote{https://github.com/FudanNLP/VCWE}. Each dataset contains a list of word pairs with a human score of how related or similar the two words are.

		We calculate the Spearman correlation \cite{spearman1904proof} between the labels and our scores generated by the embeddings. The Spearman correlation is a rank-based correlation measure that assesses how well the scores describe the true labels. The evaluation results of our model and baseline methods on word similarity datasets are shown in Table \ref{table_1}.
		
		From the results, we can see that VCWE outperforms other baseline models. The effect of CBOW is much better than Skip-gram. The impact of GWE and CWE are relatively close. The JWE model works better than other benchmark models. In the VCWE model, when we remove the CNN and the image information, the result falls by 2.91. When we replace Bi-LSTM network and self-attention with the averaging operation, the result drops by 1.42.
		
		In the last subsection, we will qualitatively analyze the results of word similarity for different models.

		\subsection{Sentiment Analysis Task}
		To evaluate the quality of our vectors regarding semantics, we use datasets\footnote{http://sentic.net/chinese-review-datasets.zip} collected by \cite{peng2018learning}, which contain Chinese reviews in four domains: notebook, car, camera, and phone. They manually labeled the sentiment polarity towards each aspect target as either positive or negative. It is a binary classification task. Similar to how we process the training data, we remove non-Chinese characters and use THULAC for performing Chinese word segmentation. We build classifiers with the bidirectional LSTM \cite{hochreiter1997long} network with self-attention \cite{lin2017structured}. We use the standard training/dev/test split and report accuracy using different embeddings generated by different methods in Table \ref{table_2}.
		
		As shown in Table \ref{table_2}, Skip-gram performs well on the combination of the four groups, but it does not perform well in the works of a particular group. JWE outstrips other baseline methods by around 1.1 points. The VCWE model has achieved outstanding results in the car, camera and phone category, with an accuracy rate of at least 3 points higher than other models, indicating that this method of training word embeddings with visual character-level features can achieve better results on downstream tasks.
		
		\subsection{Named Entity Recognition Task}
		We evaluate our model on the named entity recognition task. We use an open source Chinese NER model to test our word embeddings on MSRA dataset\footnote{https://github.com/bamtercelboo/pytorch\_NER\_PosTag\_Bi\\LSTM\_CRF}. MSRA is a dataset for simplified Chinese NER. It comes from SIGHAN 2006 shared task for Chinese NER \cite{levow2006third}. We pre-train word embeddings from different models and feed them into the input layer as features.
		
		The key to the task is to extract named entities and their associated types. Better word embeddings could get a higher F1 score of NER. The results in Table \ref{table_3} show that our model also outperforms baseline models in this task. The performance of CWE and GWE models are similar, both slightly lower than Skip-gram and CBOW models. The F1 score of the JWE model exceeds that of other baseline models and is similar to our model. When removing the CNN and image information, our LSTM with the self-attention model can also achieve the best results on this task, indicating that the learned inter-character composition is practical.
		
		\begin{table}[h!]\setlength{\tabcolsep}{5pt}
			\centering
			\begin{tabular}{l|c c c|c}
				\toprule
				\multirow{2}*{Model} & \multicolumn{3}{c|}{NER} & POS Tag\\
				\cline{2-5}
				~ & Prec. & Recall &  F1 & Acc\\
				\midrule
				Skip-gram & 85.30 & 84.18 & 84.74 & 95.87 \\
				CBOW  & 85.64 & 82.98 & 84.29 & 95.79\\
				CWE & 83.89 & 82.57 & 83.23 & 95.45\\
				GWE  & 84.06 & 82.52 & 83.28 & 95.45\\
				JWE & 85.74 & 84.87 & 85.30 & 95.91\\
				\midrule
				VCWE & 86.93 & 84.64 & \textbf{85.77} & \textbf{96.00}\\
				\quad -CNN & 86.73 & 84.83 & \textbf{85.77} & 95.92\\
				\quad -LSTM & 85.98 & 84.53 & 85.25 & 95.96\\
				\bottomrule
			\end{tabular}
			\caption{\label{table_3} Chinese NER and POS tagging results for different pretrained embeddings. The configurations are the same of the ones used in Table \ref{table_1}.
			}
		\end{table}
		
		\subsection{Part-of-speech Tagging Task}
		The evaluation is performed on the PKU's People's Daily \footnote{http://klcl.pku.edu.cn/zygx/zyxz/index.htm} (PPD) \cite{yu2001processing} with the standard training/dev/test split. The model is trained with the bidirectional LSTM model using the same hyper-parameters. Results on the POS accuracy on the test set are reported in Table \ref{table_3}.
		
		The gap between the usage of different embeddings is not significant, and our model has achieved the best results with a slight advantage.
		
		\begin{table*}[t!]\small
			\centering
			\begin{tabular}{c|c|c|c|c}
				\toprule
				Targets & Skip-gram & GWE & JWE & VCWE \\
				\midrule
				\multirow{10}*{\tabincell{c}{唐诗(Tang poetry)}} & 散曲(Qu-Poetry) &  宋诗(Song poetry) & 诗话(notes on poetry) & 诗话(notes on poetry)\\
				~ & 琴谱(music score) & 赋诗(indite) & 古今(ancient and modern) & 宋诗(Song poetry)\\
				~ & 佚(anonymity) &  诗韵(rhyme) & 佚(anonymity) & 绝句(jueju)\\
				~ & 宋词(Song Ci Poetry) &  汉诗(Chinese poetry) & 乐府(Yuefu) & 宋词(Song Ci Poetry)\\
				~ & 白居易(Bai Juyi) & 唐璜(Don Juan) & 琴谱(music score)  & 吟咏(chant)\\
				~ & 绝句(jueju) & 吟诗(recite poems) & 辑录(compile) & 乐府(Yuefu)\\
				~ & 著录(record) & 唐寅(Tang Yin) & 刻本(carving copy) & 七言(seven-character)\\
				~ & 楚辞(Chu Songs) & 唐僧(Monk Tang) & 传世(be handed down) & 李商隐(Li Shangyin)\\
				~ & 乐府(Yuefu) & 唐括(Tang Ku) & 古诗(ancient poetry) & 古诗(ancient poetry)\\
				~ & 辑录(compile) & 诗(poetry) & 散曲(Qu-Poetry)  & 诗文(poetic prose)\\
				\midrule
				\multirow{10}*{沙发(sofa)} & 办公桌(bureau) & 沙漏(hourglass) & 桌子(desk) & 衣柜(wardrobe)\\
				~ & 卧室(bedroom) & 沙尘(sand) & 衣柜(wardrobe) & 卧室(bedroom)\\
				~ & 椅子(chair) & 沙袋(sandbag) & 毛巾(washcloth) & 浴缸(bathtub)\\
				~ & 楼上(upstairs) & 沙盒(sandbox) & 书桌(secretaire) & 客厅(living room)\\
				~ & 客厅(living room) & 沙哑(raucity) & 棉被(quilt) & 窗帘(curtain)\\
				~ & 浴缸(bathtub) & 沙嘴(sandspit) & 长椅(bench) & 椅子(chair)\\
				~ & 楼下(downstairs) & 沙嗲(satay) & 窗帘(curtain) & 壁炉(fireplace)\\
				~ & 雨衣(raincoat) & 沙包(sandbag) & 浴缸(bathtub) & 房门(door)\\
				~ & 血迹(bloodstain) & 沙织(Saori) & 房门(door) & 长椅(bench)\\
				~ & 电话亭(telephone box) & 沙蚕(nereid) & 广告牌(billboard) & 桌子(desk)\\
				\bottomrule
			\end{tabular}
			\caption{\label{table_4}Case study for qualitative analysis. Given the target word, we list the top 10 similar words from each algorithm so as to observe the differences. }
		\end{table*}

		\subsection{Qualitative analysis}
		To better understand the quality of the learning word embedding for each model, we conduct a qualitative analysis by doing some case studies in Table \ref{table_4} to illustrate the most similar words for certain target words under different methods. Explicitly, we present the top 10 words that are most similar to our target word. The similar words are retrieved based on the cosine similarity calculated using the learned embeddings.
		
		The first example word we consider is ``唐诗(Tang poetry)''. It refers to poetry written in or around the time of or in the characteristic style of China's Tang Dynasty.
		
		All the top-ranked words identified by GWE contain the character ``唐(Tang)'' and ``诗(poetry)'', but in addition to the Tang Dynasty, ``唐(Tang)'' also has other meanings such as surnames. GWE yields several words such as ``唐璜(Don Juan)'', ``唐寅(Tang Yin)'', ``唐僧(Monk Tang)'' and ``唐括(Tang Ku)'', which do not appear to be semantically close to the target word. In Skip-gram and JWE, certain words such as ``佚(anonymity)'' and ``古今(ancient and modern)'' do not appear to be semantically very closely related to the target word. In our VCWE model, all the top-ranked words are semantically related to the target word, including the genre of poetry, poets of the Tang Dynasty, and so on.

		We choose the ``沙发(sofa)'' as the second target word. Like the first two words, GWE only pays attention to the character ``沙(sand)''. Skip-gram and JWE have some irrelevant words such as ``电话亭(telephone box)'' and ``广告牌(billboard)''. VCWE pays more attention to the non-compositionality, and the results are better than other models.
		
		Limited to the width of the table, we do not show the results of CWE model. The results of the GWE model are not much different from the CWE model, indicating that the image features obtained by pre-training of GWE may not play a decisive role. However, our model does not pre-train image information, but jointly trains and dynamically updates image feature information and it works better. JWE model is similar to Skip-gram model in that they pay more attention to contextual information, but sometimes the model gets some irrelevant words.
		
		\section{Discussion}
		Unlike phonograms, logograms have word and phrase meanings singularly. The images of Chinese characters contain rich semantic information. Since logographic languages are more closely associated with images than alphabet languages, it makes sense to mine the characteristics of these images.
		
		\citeauthor{liu2017learning} \shortcite{liu2017learning} provide a new way to automatically extract character-level features, creating an image for the character and running it through a convolutional neural network to produce a visual character embedding. However, this method does not utilize the rich semantic information of contextual words. Our model extracts both image features and contextual semantic information.
		
		\citeauthor{su2017learning} \shortcite{su2017learning} introduce a pixel-based model that learns character features from font images. However, they use convolutional auto-encoder(convAE) to extract image features in advance, and then add these features to the CWE \cite{chen2015joint} model. In the end, the effect of the model is not much different from CWE. Our model is an end-to-end model. We update the image's feature parameters in real time during training, and our model achieves better results than the GWE model.
		
		Our research focuses on simplified Chinese word embeddings, and the idea can also be applied to other languages that share a similar writing system, such as traditional Chinese, Japanese, and so on.
		
		\section{Conclusion and Future Work}
		In this paper, we proposed a pixel-based model to learn Chinese word embeddings with character embeddings that are compositional in the components of the characters. We utilized the visual features of Chinese characters to enhance the word embedding. We showed that our model outperforms the baseline model in the word similarity, sentiment analysis, named entity recognition and part-of-speech tagging tasks.
		
		In summary, we optimized our pixel-based word embedding method to make the model end-to-end and make full use of the contextual information. In the future, we hope to apply our model to other downstream tasks and other logographic writing systems.
		
		\section*{Acknowledgments}
		We would like to thank the anonymous reviewers for their valuable comments. The research work is supported by Shanghai Municipal Science and Technology Commission (No. 17JC1404100 and 16JC1420401),
		National Key Research and Development Program of China (No. 2017YFB1002104),
		and National Natural Science Foundation of China (No. 61672162 and 61751201).

		\bibliography{sunnysc_joint}

\begin{thebibliography}{32}
\expandafter\ifx\csname natexlab\endcsname\relax\def\natexlab#1{#1}\fi

\bibitem[{Bengio et~al.(2003)Bengio, Ducharme, Vincent, and
  Jauvin}]{bengio2003neural}
Yoshua Bengio, R{\'e}jean Ducharme, Pascal Vincent, and Christian Jauvin. 2003.
\newblock A neural probabilistic language model.
\newblock \emph{Journal of machine learning research}, 3(Feb):1137--1155.

\bibitem[{Bojanowski et~al.(2016)Bojanowski, Grave, Joulin, and
  Mikolov}]{bojanowski2016enriching}
Piotr Bojanowski, Edouard Grave, Armand Joulin, and Tomas Mikolov. 2016.
\newblock Enriching word vectors with subword information.
\newblock \emph{arXiv preprint arXiv:1607.04606}.

\bibitem[{Cao and Rei(2016)}]{cao2016joint}
Kris Cao and Marek Rei. 2016.
\newblock A joint model for word embedding and word morphology.
\newblock \emph{arXiv preprint arXiv:1606.02601}.

\bibitem[{Cao et~al.(2018)Cao, Lu, Zhou, and Li}]{cao2018cw2vec}
Shaosheng Cao, Wei Lu, Jun Zhou, and Xiaolong Li. 2018.
\newblock cw2vec: Learning chinese word embeddings with stroke n-gram
  information.

\bibitem[{Chen et~al.(2015)Chen, Xu, Liu, Sun, and Luan}]{chen2015joint}
Xinxiong Chen, Lei Xu, Zhiyuan Liu, Maosong Sun, and Huan-Bo Luan. 2015.
\newblock Joint learning of character and word embeddings.
\newblock In \emph{IJCAI}, pages 1236--1242.

\bibitem[{Harris(1954)}]{harris1954distributional}
Zellig~S Harris. 1954.
\newblock Distributional structure.
\newblock \emph{Word}, 10(2-3):146--162.

\bibitem[{Hochreiter and Schmidhuber(1997)}]{hochreiter1997long}
Sepp Hochreiter and J{\"u}rgen Schmidhuber. 1997.
\newblock Long short-term memory.
\newblock \emph{Neural computation}, 9(8):1735--1780.

\bibitem[{Kingma and Ba(2014)}]{kingma2014adam}
Diederik~P Kingma and Jimmy Ba. 2014.
\newblock Adam: A method for stochastic optimization.
\newblock \emph{arXiv preprint arXiv:1412.6980}.

\bibitem[{LeCun et~al.(1995)LeCun, Bengio et~al.}]{lecun1995convolutional}
Yann LeCun, Yoshua Bengio, et~al. 1995.
\newblock Convolutional networks for images, speech, and time series.
\newblock \emph{The handbook of brain theory and neural networks},
  3361(10):1995.

\bibitem[{Levow(2006)}]{levow2006third}
Gina-Anne Levow. 2006.
\newblock The third international chinese language processing bakeoff: Word
  segmentation and named entity recognition.
\newblock In \emph{Proceedings of the Fifth SIGHAN Workshop on Chinese Language
  Processing}, pages 108--117.

\bibitem[{Levy and Goldberg(2014)}]{levy2014neural}
Omer Levy and Yoav Goldberg. 2014.
\newblock Neural word embedding as implicit matrix factorization.
\newblock In \emph{Advances in neural information processing systems}, pages
  2177--2185.

\bibitem[{Lin et~al.(2017)Lin, Feng, Santos, Yu, Xiang, Zhou, and
  Bengio}]{lin2017structured}
Zhouhan Lin, Minwei Feng, Cicero Nogueira~dos Santos, Mo~Yu, Bing Xiang, Bowen
  Zhou, and Yoshua Bengio. 2017.
\newblock A structured self-attentive sentence embedding.
\newblock \emph{arXiv preprint arXiv:1703.03130}.

\bibitem[{Liu et~al.(2017)Liu, Lu, Lo, and Neubig}]{liu2017learning}
Frederick Liu, Han Lu, Chieh Lo, and Graham Neubig. 2017.
\newblock Learning character-level compositionality with visual features.
\newblock \emph{arXiv preprint arXiv:1704.04859}.

\bibitem[{Luong et~al.(2013)Luong, Socher, and Manning}]{luong2013better}
Thang Luong, Richard Socher, and Christopher Manning. 2013.
\newblock Better word representations with recursive neural networks for
  morphology.
\newblock In \emph{Proceedings of the Seventeenth Conference on Computational
  Natural Language Learning}, pages 104--113.

\bibitem[{Mikolov et~al.(2013{\natexlab{a}})Mikolov, Chen, Corrado, and
  Dean}]{mikolov2013efficient}
Tomas Mikolov, Kai Chen, Greg Corrado, and Jeffrey Dean. 2013{\natexlab{a}}.
\newblock Efficient estimation of word representations in vector space.
\newblock \emph{arXiv preprint arXiv:1301.3781}.

\bibitem[{Mikolov et~al.(2010)Mikolov, Karafi{\'a}t, Burget,
  {\v{C}}ernock{\`y}, and Khudanpur}]{mikolov2010recurrent}
Tom{\'a}{\v{s}} Mikolov, Martin Karafi{\'a}t, Luk{\'a}{\v{s}} Burget, Jan
  {\v{C}}ernock{\`y}, and Sanjeev Khudanpur. 2010.
\newblock Recurrent neural network based language model.
\newblock In \emph{Eleventh Annual Conference of the International Speech
  Communication Association}.

\bibitem[{Mikolov et~al.(2013{\natexlab{b}})Mikolov, Sutskever, Chen, Corrado,
  and Dean}]{mikolov2013distributed}
Tomas Mikolov, Ilya Sutskever, Kai Chen, Greg~S Corrado, and Jeff Dean.
  2013{\natexlab{b}}.
\newblock Distributed representations of words and phrases and their
  compositionality.
\newblock In \emph{Advances in neural information processing systems}, pages
  3111--3119.

\bibitem[{Miller and Charles(1991)}]{miller1991contextual}
George~A Miller and Walter~G Charles. 1991.
\newblock Contextual correlates of semantic similarity.
\newblock \emph{Language and cognitive processes}, 6(1):1--28.

\bibitem[{Peng et~al.(2018)Peng, Ma, Li, and Cambria}]{peng2018learning}
Haiyun Peng, Yukun Ma, Yang Li, and Erik Cambria. 2018.
\newblock Learning multi-grained aspect target sequence for chinese sentiment
  analysis.
\newblock \emph{Knowledge-Based Systems}, 148:167--176.

\bibitem[{Pennington et~al.(2014)Pennington, Socher, and
  Manning}]{pennington2014glove}
Jeffrey Pennington, Richard Socher, and Christopher Manning. 2014.
\newblock Glove: Global vectors for word representation.
\newblock In \emph{Proceedings of the 2014 conference on empirical methods in
  natural language processing (EMNLP)}, pages 1532--1543.

\bibitem[{Qiu et~al.(2014)Qiu, Cui, Bian, Gao, and Liu}]{qiu2014co}
Siyu Qiu, Qing Cui, Jiang Bian, Bin Gao, and Tie-Yan Liu. 2014.
\newblock Co-learning of word representations and morpheme representations.
\newblock In \emph{Proceedings of COLING 2014, the 25th International
  Conference on Computational Linguistics: Technical Papers}, pages 141--150.

\bibitem[{Rubenstein and Goodenough(1965)}]{rubenstein1965contextual}
Herbert Rubenstein and John~B Goodenough. 1965.
\newblock Contextual correlates of synonymy.
\newblock \emph{Communications of the ACM}, 8(10):627--633.

\bibitem[{Shi et~al.(2015)Shi, Zhai, Yang, Xie, and Liu}]{shi2015radical}
Xinlei Shi, Junjie Zhai, Xudong Yang, Zehua Xie, and Chao Liu. 2015.
\newblock Radical embedding: Delving deeper to chinese radicals.
\newblock In \emph{Proceedings of the 53rd Annual Meeting of the Association
  for Computational Linguistics and the 7th International Joint Conference on
  Natural Language Processing (Volume 2: Short Papers)}, volume~2, pages
  594--598.

\bibitem[{Spearman(1904)}]{spearman1904proof}
Charles Spearman. 1904.
\newblock The proof and measurement of association between two things.
\newblock \emph{The American journal of psychology}, 15(1):72--101.

\bibitem[{Su and Lee(2017)}]{su2017learning}
Tzu-Ray Su and Hung-Yi Lee. 2017.
\newblock Learning chinese word representations from glyphs of characters.
\newblock \emph{arXiv preprint arXiv:1708.04755}.

\bibitem[{Sun et~al.(2016{\natexlab{a}})Sun, Guo, Lan, Xu, and
  Cheng}]{sun2016inside}
Fei Sun, Jiafeng Guo, Yanyan Lan, Jun Xu, and Xueqi Cheng. 2016{\natexlab{a}}.
\newblock Inside out: Two jointly predictive models for word representations
  and phrase representations.
\newblock In \emph{AAAI}, pages 2821--2827.

\bibitem[{Sun et~al.(2016{\natexlab{b}})Sun, Chen, Zhang, Guo, and
  Liu}]{sun2016thulac}
Maosong Sun, Xinxiong Chen, Kaixu Zhang, Zhipeng Guo, and Zhiyuan Liu.
  2016{\natexlab{b}}.
\newblock Thulac: An efficient lexical analyzer for chinese.
\newblock Technical report, Technical Report.

\bibitem[{Wieting et~al.(2016)Wieting, Bansal, Gimpel, and
  Livescu}]{wieting2016charagram}
John Wieting, Mohit Bansal, Kevin Gimpel, and Karen Livescu. 2016.
\newblock Charagram: Embedding words and sentences via character n-grams.
\newblock \emph{arXiv preprint arXiv:1607.02789}.

\bibitem[{Xu et~al.(2016)Xu, Liu, Zhang, Li, and Chen}]{xu2016improve}
Jian Xu, Jiawei Liu, Liangang Zhang, Zhengyu Li, and Huanhuan Chen. 2016.
\newblock Improve chinese word embeddings by exploiting internal structure.
\newblock In \emph{Proceedings of the 2016 Conference of the North American
  Chapter of the Association for Computational Linguistics: Human Language
  Technologies}, pages 1041--1050.

\bibitem[{Yin et~al.(2016)Yin, Wang, Li, Li, and Wang}]{yin2016multi}
Rongchao Yin, Quan Wang, Peng Li, Rui Li, and Bin Wang. 2016.
\newblock Multi-granularity chinese word embedding.
\newblock In \emph{Proceedings of the 2016 Conference on Empirical Methods in
  Natural Language Processing}, pages 981--986.

\bibitem[{Yu et~al.(2017)Yu, Jian, Xin, and Song}]{yu2017joint}
Jinxing Yu, Xun Jian, Hao Xin, and Yangqiu Song. 2017.
\newblock Joint embeddings of chinese words, characters, and fine-grained
  subcharacter components.
\newblock In \emph{Proceedings of the 2017 Conference on Empirical Methods in
  Natural Language Processing}, pages 286--291.

\bibitem[{Yu et~al.(2001)Yu, Lu, Zhu, Duan, Kang, Sun, Wang, Zhao, and
  Zhan}]{yu2001processing}
Shiwen Yu, Jianming Lu, Xuefeng Zhu, Huiming Duan, Shiyong Kang, Honglin Sun,
  Hui Wang, Qiang Zhao, and Weidong Zhan. 2001.
\newblock Processing norms of modern chinese corpus.
\newblock Technical report, Technical report.

\end{thebibliography}
		\bibliographystyle{acl_natbib}
	\end{CJK*}
\end{document}